\documentclass[sigconf]{acmart}

\usepackage{url}
\usepackage{multirow}
\usepackage{makecell}
\usepackage{colortbl}
\usepackage{xcolor}
\usepackage{arydshln} 
\usepackage{booktabs}
\usepackage{hyperref}
\usepackage{tabularx}

\usepackage[linesnumbered,ruled,vlined]{algorithm2e}
\usepackage{algpseudocode}
\usepackage{amsmath}
\usepackage{siunitx}
\AtBeginDocument{%
  }

\copyrightyear{2025}
\acmYear{2025}
\setcopyright{acmlicensed}
\acmConference[MM '25] {Proceedings of the 33rd ACM International Conference on Multimedia}{October 27--31, 2025}{Dublin, Ireland.}
\acmBooktitle{Proceedings of the 33rd ACM International Conference on Multimedia (MM '25), October 27--31, 2025, Dublin, Ireland}
\acmDOI{10.1145/3746027.3755534}
\acmISBN{979-8-4007-2035-2/2025/10}

\acmSubmissionID{4314}

\begin{document}

\title{A Multimodal Deviation Perceiving Framework for Weakly-Supervised Temporal Forgery Localization}
\author{Wenbo Xu}
\email{xuwb25@mail2.sysu.edu.cn}
\orcid{0009-0007-8702-4521}
\affiliation{
  \institution{School of Computer Science and Engineering, Sun Yat-sen University}
  \city{Guangzhou}
  \country{China}
}

\author{Junyan Wu}
\email{wujy298@mail2.sysu.edu.cn}
\orcid{0000-0003-2692-5928}
\affiliation{
  \institution{School of Computer Science and Engineering, Sun Yat-sen University}
  \city{Guangzhou}
  \country{China}
}

\author{Wei Lu}
\orcid{0000-0002-4068-1766}
\authornote{Corresponding author.}
\email{luwei3@mail.sysu.edu.cn}
\affiliation{
  \institution{School of Computer Science and Engineering, Sun Yat-sen University}
  \city{Guangzhou}
  \country{China}
  }

\author{Xiangyang Luo}
\orcid{0000-0003-3225-4649}
\email{luoxy\_ieu@sina.com}
\affiliation{
  \institution{State Key Laboratory of Mathematical Engineering and Advanced Computing, Zhengzhou, China}
  \city{}
  \country{}
} 

\author{Qian Wang}
\orcid{0000-0002-8967-8525}
\email{qianwang@whu.edu.cn}
\affiliation{
  \institution{School of Cyber Science and Engineering, Wuhan University}
  \department{}
  \city{Wuhan}
  \country{China}
}


\begin{abstract}
Current researches on Deepfake forensics often treat detection as a classification task or temporal forgery localization problem, which are usually restrictive, time-consuming, and challenging to scale for large datasets.
To resolve these issues, we present a multimodal deviation perceiving framework for weakly-supervised temporal forgery localization (MDP), which aims to identify temporal partial forged segments using only video-level annotations.
The MDP proposes a novel multimodal interaction mechanism (MI) and an extensible deviation perceiving loss to perceive multimodal deviation, which achieves the refined start and end timestamps localization of forged segments.
Specifically, MI introduces a temporal property preserving cross-modal attention to measure the relevance between the visual and audio modalities in the probabilistic embedding space.
It could identify the inter-modality deviation and construct comprehensive video features for temporal forgery localization.
To explore further temporal deviation for weakly-supervised learning, an extensible deviation perceiving loss has been proposed, aiming at enlarging the deviation of adjacent segments of the forged samples and reducing that of genuine samples. 
Extensive experiments demonstrate the effectiveness of the proposed framework and achieve comparable results to fully-supervised approaches in several evaluation metrics.
\end{abstract}



\settopmatter{printacmref=true} 


\maketitle
\section{Introduction}
Generative artificial intelligence has rapidly advanced in recent years, utilizing existing Artificial Intelligence Generated Content (AIGC) technology could generate high-quality multimedia content such as image, audio, video, etc.
Deepfake, as a specific application of AIGC technology, allows for manipulating multimedia content of actual people or generating fictional content.  
However, the misuse of Deepfake represents a substantial threat to individual privacy, copyright protection, and the overall stability of society.

\begin{figure}[t]
    \centering
    \includegraphics[width=\linewidth]{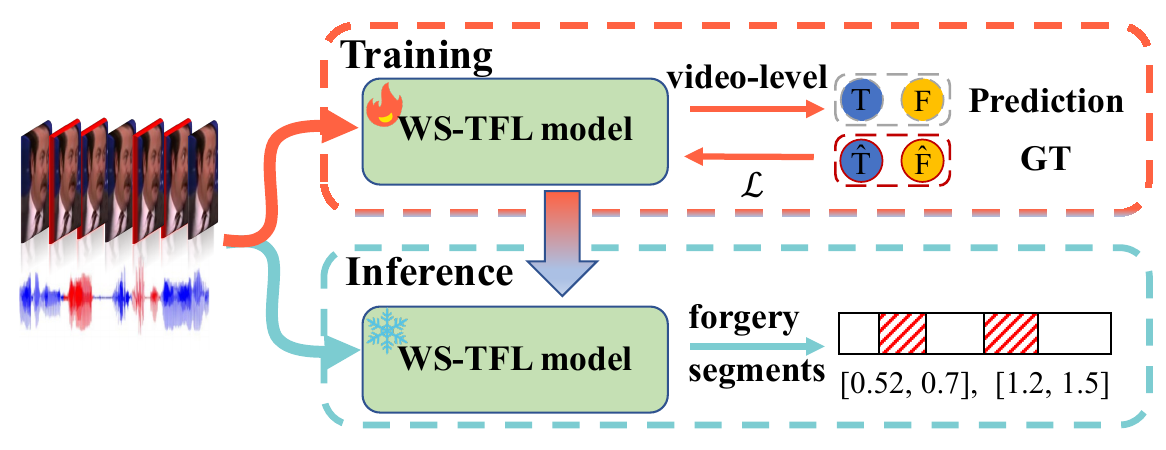}
    \caption{The schematic diagram of weakly-supervised temporal forgery localization task (WS-TFL).
    In the training phase, merely video-level fake (F) and true (T) annotations are utilized for loss calculation and model parameter updating.
    In the inference phase, for a given video, the timestamps of forged segments are predicted with the trained model.}
    \label{fig:WS-TFL}
\end{figure}

Current research in Deepfake forensics primarily tackles the issue through classification tasks, particularly binary classification for videos or images \cite{yin2023dynamic,gu2022region,dong2023implicit}.
Nevertheless, this methodology exhibits limitations when addressing more challenging deepfake scenarios, particularly in the context of temporal partial forgery localization.
Considering the specificity and potential pernicious effects of temporal partial forgery, Chugh ~\shortcite{chugh2020not} proposed the temporal forgery localization task (TFL) to localize the start and end timestamps of forged segments.
Several researches \cite{cai2023glitch,liu2023audio} have explored the TFL task with a fully-supervised methodology.
Both BA-TFD+ \cite{cai2023glitch} and AVTFD \cite{liu2023audio} attempted to combine the TFL with frame-level Deepfake detection methods.
UMMAFormer \cite{zhang2023ummaformer} aimed to mine forgery traces through feature reconstruction.
The aforementioned fully-supervised temporal forgery localization (FS-TFL)  methods have achieved some degree of localization performance.
However, they require elaborate frame-level or timestamp annotations for fully-supervised learning, which is usually costly and time-consuming.

To cope with the dilemma of FS-TFL, weakly-supervised learning is introduced to TFL.
The schematic diagram of weakly-supervised temporal forgery localization (WS-TFL) is shown in Figure~\ref{fig:WS-TFL}.
The main challenges of WS-TFL are: 1) integrating multimodal information between visual and audio features, 
and 2) leveraging video-level annotations to mine subtle forgery traces for temporal partial forgery localization.
The weakly-supervised learning allows training on imprecise, partially accurate, or noisy annotations, enabling more refined inference tasks \cite{zhou2018brief}.
The existing weakly-supervised learning methods are mainly for computer vision tasks with strong semantic signals like temporal action localization \cite{yun2024weakly} and object detection \cite{zhang2021weakly}, and focus primarily on the single visual modality.
Therefore they are inappropriate for tracing subtle forgery traces in multimodal Deepfake scenarios \cite{zhang2023ummaformer}.

To overcome these challenges, we present a multimodal deviation perceiving framework for weakly-supervised temporal forgery localization (MDP) in this paper, which aims to identify the timestamps of temporal partial forged segments using only video-level annotations.
A novel multimodal interaction mechanism (MI) is introduced to analyze the dissimilarity or inter-modality deviation between visual and audio features.
MI utilizes a temporal property preserving cross-modal attention to integrate multimodal information and constructs comprehensive video features for temporal forgery localization.
Besides, we propose an extensible deviation perceiving loss to explore further temporal deviation for weakly-supervised learning, which explores further temporal deviation by measuring the degree of deviation between adjacent segments.

Specifically, the present framework consists of three modules: feature extraction, multimodal interaction, and temporal forgery localization.
Feature extraction module first extracts visual and audio features of a given video using pre-trained models.
The visual and audio modalities are regarded as distinct encoding formats with relevance.
The multimodal interaction module transforms the visual and audio features into token space and aligns them in temporal and spatial dimensions.
A temporal property preserving cross-modal attention is utilized to enhance the multimodal features, thereby generating comprehensive video features by concatenating all the visual and audio features.
Finally, the temporal forgery localization module generates a temporal forgery activation sequence (FAS) based on the comprehensive video features.
In the training phase, the video-level prediction is obtained by summing the FAS for weakly-supervised learning.
While in the inference phase, the start and end timestamps of the forged segments are obtained according to the FAS.
Moreover, an extensible deviation perceiving loss is proposed to measure the degree of deviation between adjacent segments.
The MDP improves the localization precision by enlarging the deviation of adjacent segments of the forged samples and reducing that of genuine samples.
The main contributions are summarized as follows:

\begin{itemize}
    \item We propose a multimodal deviation perceiving framework for weakly-supervised temporal forgery localization, which could identify the timestamps of temporal forged segments using only video-level annotations. 
    \item A temporal property preserving cross-modal attention is proposed, which is to perceive the inter-modality deviation between the visual and audio features and construct representative comprehensive video features.
    \item An extensible deviation perceiving loss is proposed for weakly-supervised learning, which aims at enlarging the temporal deviation of forged samples while reducing that of genuine samples.
    \item Extensive experiments have been conducted on two challenging datasets to demonstrate the effectiveness of the proposed framework, and MDP achieves comparable results to fully-supervised approaches in several evaluation metrics.

\end{itemize}

\section{Related Work}
\subsection{Multi-modal Deepfake Detection}
With the gradual progression of Deepfake forensics research, research on multimodal approach utilizing both visual and audio information is becoming increasingly popular \cite{jia2024can,xia2024mmnet,chen2023jointly,nirkin2021deepfake}.
The primary issue in multimodal Deepfake detection is to identify forgery traces from two distinct embedding spaces.
Chugh ~\shortcite{chugh2020not} and McGurk ~\shortcite{mcgurk1976hearing} extracted the visual and audio features and compared the discrepancies between the two modalities directly.
To fully facilitate the fusion of multimodal features, Zhou and Lim ~\shortcite{zhou2021joint} conducted joint audio-visual learning to promote the interaction between visual and audio modalities.
Meanwhile, Yin ~\shortcite{yin2024fine} analyzed the relationships of intra- and inter-modality by the heterogeneous graph and achieved the fine-grained multimodal Deepfake classification target.
To tackle the temporal partial forgery localization challenge \cite{cai2022you}, Zhang ~\shortcite{zhang2023ummaformer} proposed to predict forged segments by multimodal feature reconstruction.
Nie \cite{nie2024frade} proposed forgery-aware audio-distilled multimodal learning by capturing high-frequency discriminative features for Deepfake detection.
As visual and audio modalities have substantial discrepancies in macro-semantics and feature distributions, they are regarded as distinct encoding formats with relevance \cite{shvetsova2022everything}.
We transform the visual and audio features into token space \cite{lim2024probabilistic}, and then align them in temporal and spatial dimensions, and then enhance the multimodal features with a temporal property preserving cross-modal attention.

\subsection{Weakly-Supervised Learning}
There are three typical types of weakly-supervised learning: incomplete supervision, where only a subset of training data is given with annotations; inexact supervision, where the training data are given with only coarse-grained annotations; and inaccurate supervision, where the given annotations are not always accurate \cite{zhou2018brief}.
Weakly-Supervised learning has achieved numerous progress in computer vision fields such as object detection \cite{fu2024cf,su2024consistency} and temporal action localization \cite{su2024consistency,li2024complete}.
In the object detection domain, weakly-supervised object localization (WSOL) and weakly-supervised object detection (WSOD) are treated as two different tasks \cite{zhang2021weakly}.
WSOL mainly aims at entailing the location of a single object utilizing merely image-level annotations \cite{chen2024adaptive}.
While the goal of WSOD is to detect every possible object with image-level annotations instead.
Weakly-supervised temporal action localization (WS-TAL) is proposed to predict the category and start-end timestamps of actions within a video, training with only video-level action category annotations \cite{wang2023temporal}.
The goals of WS-TAL and WS-TFL are analogous.
However, WS-TAL approaches primarily focus on the semantic perception of the video and target visual modality.
Hence it is inappropriate for multimodal Deepfake scenarios that require weak signal perception as forgery traces \cite{zhang2023ummaformer,sheng2024dirloc,sheng2025sumi,wu2025weakly}. 

\section{Multimodal Deviation Perceiving Framework}

\begin{figure*}[htbp]  
    \centering
    \includegraphics[width=\textwidth]{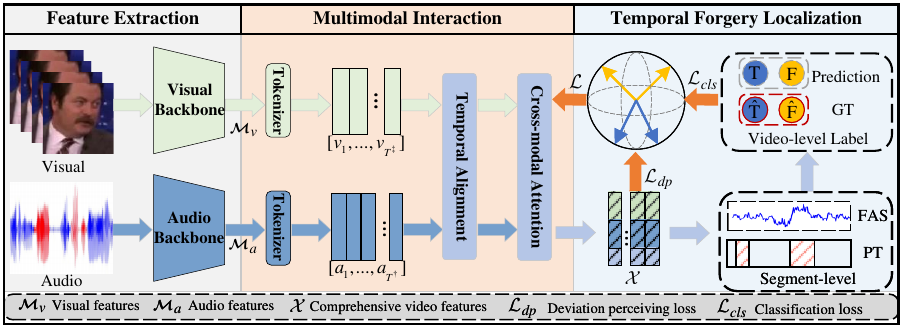}  
    \caption{Diagrammatic overview of the proposed multimodal deviation perceiving framework for weakly-supervised temporal forgery localization. GT denotes the video-level ground truth annotation. 
Prediction denotes the video-level prediction result for weakly-supervised learning.
FAS is the temporal forgery activation sequence obtained from comprehensive video features.
And PT is the predicted timestamps of forged segments.}
    \label{fig:MDP_structure}
\end{figure*}

\subsection{Problem Definition}
The WS-TFL aims to localize the timestamps of all forged segments in Deepfake video, depending solely on the video-level annotations.
Specifically, given a set of videos with video-label annotations available $\mathcal{D}=\left\{v_i, y_i\right\}_{i=1}^N$, where $y_i \in \left\{0, 1\right\}$ represents the video $v_i$ is genuine or forged, $N$ is the total number of videos. 
In the training phase, merely $\mathcal{Y} = \left\{y_i\right\}_{i=1}^N$ are accessible for loss calculation and model parameter learning under supervised paradigm.
During the inference phase, the WS-TFL model should predict all the forged segments $\mathcal{F}=\left\{s_j, e_j\right\}_{j=1}^K$ of a given video $v$, where $s_j$ and $e_j$ indicate the start and end timestamp of the $j$-th forged segment, and $K$ is the total number of forged segments in $v$.

\subsection{Overview}
To establish a universal framework to facilitate the research and development of WS-TFL tasks, we propose a multimodal deviation perceiving framework for weakly-supervised temporal forgery localization, as shown in Figure ~\ref{fig:MDP_structure}, which aims to identify temporal forged segments using merely video-level annotations.
We mine the multimodal deviation for temporal forgery localization under the supervision of weak video-level annotations.
The inter-modality deviation between visual and audio features is obtained by multimodal interaction with temporal property preservation.
Additionally, we investigate the temporal deviation between adjacent segments using a deviation perceiving loss.

Specifically, given an arbitrary video dataset $\mathcal{D}=\left\{v_i, y_i\right\}_{i=1}^N$, which merely video-level annotations are accessible, the pre-trained feature extractors (\textit{e.g.}, TSN \cite{wang2018temporal} or ResNet \cite{he2016deep} for visual modality, and BYOL-A \cite{niizumi2021byol} or Wav2Vec \cite{baevski2020wav2vec} for audio modality) are first utilized to extract corresponding visual modality frame-level features $\mathcal{M}_v = \left\{v_t \in \mathbb{R}^{h^\ddagger \times w^\ddagger}\right\}_{t=1}^{T^{\ddagger}}$ and audio modality frame-level features $\mathcal{M}_a = \left\{a_t \in \mathbb{R}^{h^\dagger \times w^\dagger} \right\}_{t=1}^{T^{\dagger}}$.
Following that, the multimodal interaction module transforms the features into token space, enabling effective alignment of multimodal features across both temporal and spatial dimensions.
Subsequently, a cross-modal attention is utilized to enhance the visual and audio features by means of temporal property preservation in probabilistic embedding space.
The comprehensive video features $\mathcal{X} = \left\{x_t \in \mathbb{R}^{d}\right\}_{t=1}^{T}$ are obtained from the multimodal interaction module by concatenating all the visual and audio features.

Then, the FAS $\mathcal{P} = \left\{p_t \in \mathbb{R}^{2} \right\}_{t=1}^T$ could be derived from $\mathcal{X}$ by utilizing a classifier, where the two variables of $p_t$ indicate the probability that the $t$-th segment is genuine or forged, respectively.
In the training phase, the video-level prediction result $\hat{y} \in \mathbb{R}^{2}$ could be derived by summing $\mathcal{P}$ for weakly-supervised learning.
\begin{equation}
\hat{y} = \sigma\left(\frac{1}{T}\sum_{t=1}^T p_t\right)
\end{equation}
where $\sigma$ is the normalization operation.
During the inference phase, the prediction results for each segment are obtained based on the $\mathcal{P}$.
Forged segments are identified depending on the genuine and forged probability of $p_t$, and consecutive forged segments are merged into the same group. 
These results are subsequently integrated with the temporal information of the video to obtain all the forged segments $\mathcal{F}=\left\{s_j, e_j\right\}_{j=1}^K$.

\subsection{Multimodal Interaction}
\label{sec:MI}
Given an untrimmed video, the crucial task of multimodal interaction is to mine the inter-modality deviation between visual and audio features for temporal forgery localization.
To achieve this purpose, comprehensive video features $\mathcal{X}$ should be constructed from visual and audio modality.
Consequently, we propose a novel multimodal interaction mechanism (MI) that consists of feature alignments and cross-modal attention.


\textbf{Feature alignment.}
The features extracted from visual and audio modalities with the pre-trained models are commonly non-aligned in temporal and spatial dimensions.
For multimodal Deepfake detection and localization,
feature alignment operations are particularly important, especially in the temporal dimension.

To address this problem for further multimodal interaction, we should align the multimodal features first.
The visual and audio features are tokenized at the frame-level in spatial dimension.
In terms of visual modality, the features are divided into $T^\ddagger$ frames along the temporal dimension.
The features of each frame are then tokenized into a feature vector $v_t \in \mathbb{R}^d$, obtained the visual modality features $\mathcal{M}_v^{\prime} = \left\{v_t \in \mathbb{R}^{d}\right\}_{t=1}^{T^{\ddagger}}$.
Similarly, the audio modality features $\mathcal{M}_a^{\prime} = \left\{a_t \in \mathbb{R}^{d}\right\}_{t=1}^{T^{\dagger}}$ could be generated.

Since the audio modality possesses a higher frequency of sampling points per unit of time than the visual modality, the number of frames in the audio modality is distinct from that in the visual modality in a video ($T^{\ddagger} \neq T^{\dagger}$).
Consequently, it is essential to ensure the alignment of temporal dimension.
\begin{equation}
         \mathcal{M}_v^{\prime\prime} = \mathbb{A}_v(\mathcal{M}_v^{\prime})
         \label{eq1}
\end{equation}
\begin{equation}
        \mathcal{M}_a^{\prime\prime} = \mathbb{A}_a(\mathcal{M}_a^{\prime}) 
\end{equation}
where $\mathbb{A}_v$ and $\mathbb{A}_a$ are two pooling operations scaling the $\mathcal{M}_v^{\prime}$ and $\mathcal{M}_a^{\prime}$ into $T$ segments along the temporal dimension.

\begin{algorithm}[t]
\SetKwInput{Para}{\textbf{Parameter}}
\caption{The algorithm of cross-modal attention}
\label{alg}
\KwIn{Visual features $\mathcal{M}_v^{\prime\prime}$, audio features $\mathcal{M}_a^{\prime\prime}$, probabilistic encoder $\mathbb{P}_v$ and $\mathbb{P}_a$, parameter $d$, learnable parameters $\mathbf{W}_q$, $\mathbf{W}_k$, $\mathbf{W}_v$. }
\KwOut{Enhanced features $\mathbb{ATT}_{v}$ and $\mathbb{ATT}_{a}$.}
Calculate the probabilistic embeddings $\widehat{\mathcal{M}_v}$, $\widehat{\mathcal{M}_a}$ in Eq. \ref{eqp1} and Eq. \ref{eqp2}\;
Calculate the $\mathbf{Q}$, $\mathbf{K}$, $\mathbf{V}$ in Eq. \ref{eqqkv}\;
Calculate the relevance matrix $\mathcal{R}=\frac{\mathbf{Q} \cdot\mathbf{K}^{\mathbb{T}}}{\sqrt{d}}$\;
\For{$t = 1 : T$}
{
    $r_{t} = \sum_{i=1}^{T}\mathcal{R}_{it}$\;
}
Normalize $\left\{r_t \right\}_{t=1}^T$, $\widehat{\mathcal{R}}= \left[r_t\right]^{1 \times T}$\;
Calculate enhanced visual features $\mathbb{ATT}_{v} = \widehat{\mathcal{R}}^{\mathbb{T}} \cdot \mathbf{V}$\;
Similarly, calculate $\mathbb{ATT}_{a}$\;
\textbf{Return} $\mathbb{ATT}_{v}$ and $\mathbb{ATT}_{a}$
\end{algorithm}

\textbf{Cross-modal attention.}
Since WS-TFL requires mining subtle forgery traces in multimodal temporal features to predict the start and end timestamps of the forged segments, it is essential to ensure temporal information is not disrupted during multimodal interaction.
A novel temporal property preserving cross-modal attention is proposed in MDP.

As mentioned above, the visual and audio modality features could be regarded as two distinct encoding formats which have different embedding spaces.
The macro semantics and feature distributions often have substantial discrepancies.
The $\mathcal{M}_v^{\prime\prime}$ and $\mathcal{M}_a^{\prime\prime}$ are converted into probabilistic embedding space \cite{chen2020simple,lim2024probabilistic} for cross-modal attention computation.
\begin{equation}
         \widehat{\mathcal{M}_v} = \mathbb{P}_v(\mathcal{M}_v^{\prime\prime})
         \label{eqp1}
\end{equation}
\begin{equation}
         \widehat{\mathcal{M}_a} = \mathbb{P}_a(\mathcal{M}_a^{\prime\prime}) 
         \label{eqp2}
\end{equation}
where $\mathbb{P}$ is obtained by a MLP with one hidden layer.
Specifically, $\mathbb{P}(\cdot) = LN(ReLu(W^{(1)}(\cdot)))$, where $LN(\cdot)$ is a LayerNorm process.

For the visual modality features $\widehat{\mathcal{M}_v} = \left\{v_t \in \mathbb{R}^{d} \right\}_{t=1}^T$ and the audio modality features $\widehat{\mathcal{M}_a} = \left\{a_t \in \mathbb{R}^{d} \right\}_{t=1}^T$, firstly calculate the relevance between each video segment $v_t$ and audio segment $a_t$.
Thus a relevance matrix could be obtained $\mathcal{R} = \left[\mathcal{R}_{{t}^{\prime}{t}^{\prime\prime}}\right]^{T \times T}$, where $\mathcal{R}_{{t}^{\prime}{t}^{\prime\prime}}$ represents the relevance of visual segment $v_{{t}^{\prime}}$ and audio segment $a_{{t}^{\prime\prime}}$.
For 2-D visual and audio modality features, the row dimension preserves the temporal information of the corresponding video, which is crucial in temporal forgery localization.
Note that if we directly calculate the dot-product of $\mathcal{R}$ and audio modality features $\widehat{\mathcal{M}_a}$, the obtained cross-modal features have already dropped the temporal information.
To preserve the temporal property, the relevance matrix $\mathcal{R}$ is summed by columns to obtain the matrix $\widehat{\mathcal{R}} = \left[r_t\right]^{1 \times T}$, where $r_t$ represents the relevance of visual modality features $\widehat{{M}_v}$ and audio segment $a_t$.
Finally, the enhanced visual features $\mathbb{ATT}_{v}$ are derived by multiplying each $r_t$ with audio segment $a_t$.
Formally,

\begin{equation}
\mathbf{Q}=\widehat{\mathcal{M}_v} \mathbf{W}_q, \quad \mathbf{K}=\widehat{\mathcal{M}_a} \mathbf{W}_k, \quad \mathbf{V}=\widehat{\mathcal{M}_a} \mathbf{W}_v \\
\label{eqqkv}
\end{equation}
\begin{equation}
\mathbb{ATT}_{v}=\sigma\left(\mathbb{S}\left(\frac{\mathbf{Q} \cdot \mathbf{K}^{\mathbb{T}}}{\sqrt{d}}\right)\right)^{\mathbb{T}}\mathbf{V}
\end{equation}
where $\mathbf{W}_q$, $\mathbf{W}_k$, $\mathbf{W}_v$ are learnable parameters, $\mathbb{S}$ indicates the column summation, $\sigma$ is the normalization operation, and $\mathbb{T}$ indicates the matrix transpose.
Likewise, the enhanced audio features $\mathbb{ATT}_{a}$ could be calculated.
The details of cross-modal attention are shown in Algorithm \ref{alg}.

\begin{equation}
\mathcal{X} = cat(\widehat{{M}_v}, \mathbb{ATT}_{v}, \widehat{{M}_a}, \mathbb{ATT}_{a})
\end{equation}

The comprehensive video features $\mathcal{X}$ are obtained by concatenating the visual modality features $\widehat{{M}_v}$, $\mathbb{ATT}_{v}$ and the audio modality features $\widehat{{M}_a}$, $\mathbb{ATT}_{a}$. 
The proposed cross-modal attention mines inter-modality deviation while preserving the temporal information of visual and audio features. 
Therefore $\mathcal{X}$ could be utilized to perceive further temporal information for WS-TFL.

\subsection{Deviation Perceiving loss}
The WS-TFL has merely video-level annotations, which makes it difficult to validly exploit temporal information for the timestamps localization of the forged segments.
Therefore, we require digging further temporal information for weakly-supervised.
Typically, video samples obtained from devices like video cameras or smartphones exhibit minimal changes in content and statistical property between adjacent frames, both visual modality and audio modality. In contrast, forged samples are often created by splicing forged frames together, and the forged frames are often obtained by a deep learning model \cite{yin2024fine}.
Maintaining content coherence among the frame-by-frame spliced forged segments is challenging.
As a result, there are often considerable deviations between adjacent forged frames, as well as between these forged frames and the genuine frames. 
Furthermore, the data generated by the deep learning model often have relative discrepancies in statistical property compared to the genuine data.

Considering that for temporal partial forgery samples, the deviation between the forged segments and the adjacent genuine segments will be larger than that of the genuine samples \cite{zhu2024deepfake,lu2023detection}.
An extensible deviation perceiving loss is proposed to explore further temporal information for weakly-supervised learning.
Specifically, given the comprehensive video features $\mathcal{X} = \left\{x_t\right\}_{t=1}^{T}$, we calculate the temporal deviation $d$ based on the deviation of adjacent segments.
Formally,
\begin{equation}
    d=\sigma\left(\sum_{t=1}^T f\left(x_t, x_{t+1}\right)\right)
\end{equation}
\begin{equation}
f(x_t, x_{t+1})= E((x_t - x_{t+1})^2)
\label{relative entropy}
\end{equation}
where $f\left(x_t, x_{t+1}\right)$ indicates the deviation between the $t$-th segment and the $(t+1)$-th segment, and $f(\cdot)$ is a deviation measure function that measures the degree of deviation (\textit{e.g.,} mean square error (MSE) as shown in Eq. \eqref{relative entropy}).
We assessed the impact of different $f(\cdot)$ on the performance of temporal forgery localization in Section \ref{DMF}.

The temporal deviation $d$ of forgery samples is commonly larger than that of the genuine samples as the perturbation of the forged segments.
The deviation perceiving loss $\mathcal{L}_{dp}$ is introduced to constrain the MDP to enlarge the temporal deviation of forgery samples while reducing that of genuine samples. 
$\mathcal{L}_{dp}$ is calculated as
\begin{equation}
    \mathcal{L}_{dp}=-\frac{1}{N} \sum_{i=1}^{\mathrm{N}}\left[\left(1-y_i\right)\log \left(1-d_i\right)+ y_i\log \left(d_i\right)\right]
\end{equation}
where $y_i$ is the video-level annotation, and $N$ is the total number of the train dataset samples.

\subsection{Training and Inference}
Given a video $v$ with merely video-level annotations accessible, the pre-trained model is first utilized to extract the visual modality features $\mathcal{M}_v$ and audio modality features $\mathcal{M}_a$, respectively.
Then as mentioned in Section \ref{sec:MI}, feature alignment and multimodal interaction are conducted on the $\mathcal{M}_v$ and $\mathcal{M}_a$ to obtain the comprehensive video features $\mathcal{X}$.
$\mathcal{X}$ is fed into the temporal forgery activation head to generate the FAS $\mathcal{P} = \left\{p_t \in \mathbb{R}^{2} \right\}_{t=1}^T$.
The video-level prediction $\hat{y} \in \mathbb{R}^{2}$ by summing $\mathcal{P}$ along the temporal $T$. 
The overall loss function is defined as follows:
\begin{equation}
    \mathcal{L} = \mathcal{L}_{cls} + \phi\mathcal{L}_{dp}
\end{equation}
where $\mathcal{L}_{cls}$ is the video-level classification loss and $\mathcal{L}_{dp}$ is the deviation perceiving loss.
$\phi$ is a hyperparameter to balance the relationship between different losses.

In the inference phase, the MDP predicts all the forged segments $\mathcal{F}=\left\{s_j, e_j\right\}_{j=1}^K$ of a given video, where $s_j$ and $e_j$ indicate the start and end timestamps of the $j$-th forged segment.

\section{Experiments}
\subsection{Experimental Setup}
\textbf{Datasets:}
We conduct experiments \footnote{The code is available at: \url{https://github.com/wenboxu98/MDP}.}  on two challenging Deepfake temporal partial forgery datasets LAV-DF \cite{cai2023glitch} and AV-Deepfake1M \cite{cai2024av}.
\textbf{LAV-DF} is a strategic content-driven multimodal forgery dataset, which contains \num{36431} genuine videos and \num{99873} forged videos. The duration of forged segments is in the range of $[0-1.6s]$.
\textbf{AV-Deepfake1M} is a large-scale multimodal forgery dataset which contains \num{2068} subjects resulting in \num{286721} genuine videos and \num{860039} forged videos.
There are four types of samples (real, visual-only forgery, audio-only forgery and audio-visual forgery) in both LAV-DF and AV-Deepfake1M.

\textbf{Baseline Methods:}
To demonstrate the effectiveness of the proposed MDP, the fully-supervised temporal localization approaches MFMS \cite{zhang2024mfms}, UMMAFormer \cite{zhang2023ummaformer}, ActionFormer \cite{zhang2022actionformer}, TriDet \cite{shi2023tridet} are chosen for comparison.
Due to the lack of current research on WS-TFL, the WS-TAL approaches CoLA \cite{zhang2021cola}, FuSTAL \cite{feng2024full} are selected for comparison.
TAL approaches primarily focus on the visual modality as a research subject.
We utilize the real and audio-visual forgery samples for model training and evaluation metrics calculations.

\textbf{Evaluation metrics:}
The average precision (AP) and average recall (AR) are utilized as the evaluation metrics following \cite{cai2023glitch} and \cite{zhang2023ummaformer}.
For LAV-DF, the IoU thresholds of AP are set as $0.5$, $0.75$ and $0.95$.
As AV-Deepfake1M is a more challenging dataset compared to LAV-DF, 
the IoU thresholds of AP are set as $[0.1:0.1:0.7]$.
AR is calculated using $20$, $10$, $5$, and $2$ proposals with IoU thresholds $[0.5:0.05:0.95]$, respectively.

\textbf{Implementation details:}
The MDP is trained by Adam optimizer with a learning rate of $1e-5$, a batch size of 32.
The hyperparameter $\phi$ is set as $0.5$.

\begin{table*}[t]
	\centering
\begin{tabularx}{\textwidth}{c|c|XXXX|XXXXX}
\hline \multirow{2}{*}{ Method } & \multirow{2}{*}{ Supervision } & \multicolumn{4}{c|}{ AP@IoU(\%) } & \multicolumn{5}{c}{ AR@Proposals(\%) } \\
& & 0.5 & 0.75 & 0.95 & Avg. & 20 & 10 & 5 & 2 & Avg. \\
\hline ActionFormer & \multirow{4}{*}{ fully} & 96.75 & 94.49 & 30.7 & 73.98 & 92.02 & 91.55 & 90.08 & 87.37 & 90.26 \\
TriDet &  & 96.18 & 92.96 & 18.98 & 69.37 & 90.33 & 89.49 & 87.69 & 84.86 & 88.09 \\
UMMAFormer & & 98.79 & 97.24 & 53.89 & 83.31 & 95.26 & 94.91 & 94.07 & 90.4 & 93.66 \\
MFMS & & 98.89 & 97.19 & 51.14 & 82.41 & 94.96 & 94.54 & 93.71 & 90.18 & 93.35 \\
\hline CoLA & \multirow{3}{*}{ weakly} & 8.79 & 4.56 & 0.03 & 4.46 & 45.29 & 41.65 & 27.07 & 5.22 & 29.81 \\
FuSTAL &  & 18.79 & 5.61 & 0.08 & 8.16 & 26.45 & 24.27 & 22.2 & 18.45 & 22.84 \\
\cellcolor[rgb]{ .906,  .902,  .902}MDP & & \cellcolor[rgb]{ .906,  .902,  .902}$\mathbf{84.57}$ & \cellcolor[rgb]{ .906,  .902,  .902}$\mathbf{75.91}$ & \cellcolor[rgb]{ .906,  .902,  .902}$\mathbf{0.58}$ & \cellcolor[rgb]{ .906,  .902,  .902}$\mathbf{53.69}$ & \cellcolor[rgb]{ .906,  .902,  .902}$\mathbf{72.85}$ & \cellcolor[rgb]{ .906,  .902,  .902}$\mathbf{72.85}$ & \cellcolor[rgb]{ .906,  .902,  .902}$\mathbf{72.63}$ & \cellcolor[rgb]{ .906,  .902,  .902}$\mathbf{69.05}$ & \cellcolor[rgb]{ .906,  .902,  .902}$\mathbf{71.85}$ \\
\hline
\end{tabularx}
	\caption{Temporal forgery localization results of both fully-supervised and weakly-supervised approaches on LAV-DF.} 
	\label{lavdf_comp}
\end{table*}

\begin{table*}[t]\Large
	\centering
    
    \renewcommand\arraystretch{1}
    \resizebox{\textwidth}{!}{
\begin{tabular}{c|c|cccccccc|ccccc}
\hline \multirow{2}{*}{ Method } & \multirow{2}{*}{ Supervision } & \multicolumn{8}{c|}{ AP@IoU(\%) } & \multicolumn{5}{c}{ AR@Proposals(\%) } \\
& &0.1&0.2&0.3&0.4 & 0.5 & 0.6 & 0.7 & Avg. & 20 & 10 & 5 & 2 & Avg. \\
\hline 
ActionFormer & \multirow{4}{*}{ fully} & 99.66 & 99.65 & 99.59 & 99.48 & 99.25 & 98.6 & 96.66 & 98.98 & 91.35 & 90.55 & 88.55 & 84.02 & 88.62 \\
TriDet &  & 96.02 & 95.66 & 95.05 & 94.34 & 93.34 & 91.39 & 86.37 & 93.17 & 82.4 & 80.72 & 78.35 & 72.86 & 78.58 \\
UMMAFormer &  & 99.84 & 99.82 & 99.78 & 99.73 & 99.53 & 99.01 & 97.5 & 99.32 & 89.96 & 89.19 & 87.94 & 84.42 & 87.88 \\
MFMS &  & 99.01 & 98.9 & 98.8 & 98.67 & 98.42 & 97.83 & 96.04 & 98.24 & 89.75 & 88.82 & 87.23 & 83.09 & 87.22 \\
\hline 
\cellcolor[rgb]{ .906,  .902,  .902}MDP & weakly & \cellcolor[rgb]{ .906,  .902,  .902}$\mathbf{90.21}$ & \cellcolor[rgb]{ .906,  .902,  .902}$\mathbf{88.45}$ & \cellcolor[rgb]{ .906,  .902,  .902}$\mathbf{76.96}$ & \cellcolor[rgb]{ .906,  .902,  .902}$\mathbf{50.81}$ & \cellcolor[rgb]{ .906,  .902,  .902}$\mathbf{22.39}$ & \cellcolor[rgb]{ .906,  .902,  .902}$\mathbf{5.21}$ & \cellcolor[rgb]{ .906,  .902,  .902}$\mathbf{0.38}$ & \cellcolor[rgb]{ .906,  .902,  .902}$\mathbf{47.77}$ & \cellcolor[rgb]{ .906,  .902,  .902}$\mathbf{10.2}$ & \cellcolor[rgb]{ .906,  .902,  .902}$\mathbf{10.17}$ & \cellcolor[rgb]{ .906,  .902,  .902}$\mathbf{9.68}$ & \cellcolor[rgb]{ .906,  .902,  .902}$\mathbf{5.38}$ & \cellcolor[rgb]{ .906,  .902,  .902}$\mathbf{8.86}$ \\
\hline
\end{tabular}
}
	\caption{Temporal forgery localization results of both fully-supervised and weakly-supervised approaches on AV-Deepfake1M. 
    CoLA and FuSTAL are not displayed because they could not localize the timestamps of forged segments effectively.} 
	\label{av1m_comp}
\end{table*}

\subsection{Performance Comparisons}
In this section, we compare the proposed MDP with previous state-of-the-art approaches on LAV-DF and AV-Deepfake1M.
The two datasets are both for Deepfake multimodal scenarios.
For LAV-DF, the TSN \cite{wang2018temporal} and Wav2Vec \cite{baevski2020wav2vec} are utilized as the visual feature extractor and the audio feature extract, respectively.
For AV-Deepfake1M, the ResNet50 \cite{he2016deep} and Wav2Vec are used as the visual feature extractor and the audio feature extract, respectively.
All comparison approaches were retrained on the pre-trained features according to the open source code in the paper.

\textit{\textbf{LAV-DF Dataset:}} As shown in Table \ref{lavdf_comp}, the results show that MDP, which is a weakly-supervised temporal forgery localization approach, achieves relatively good performance on both AP and AR.
Compared to the weakly-supervised temporal action localization approaches, the MDP is significantly improved in both AP and AR.
It could be found that the AR@$2$ to AR@$20$ remain consistent, which indicates the proposed framework could predict the forged segments with less number of candidate proposals.
As for the AP evaluation metric, MDP achieves significantly superior performance compared to the comparison weakly-supervised approaches on both AP@$0.5$ and AP@$0.75$, which indicates that MDP exhibits higher accuracy in predicting forged segments.
It should be noted that despite MDP showing satisfactory performance on most evaluation metrics, it underperforms on the AP@$0.95$, implying that it is inadequate in localizing the precise timestamps of the forged segments.

Obviously, compared to the weakly-supervised approaches, the fully-supervised approaches achieve superior performance on both AP and AR.
Such results are also reasonable, as fully-supervised approaches are more adept at learning the relationship between the Deepfake video features and corresponding timestamps of forged segments with provided frame-level annotations.
Nevertheless, it could be observed that the MDP also achieves relatively good performance on AP@$0.5$ and AP@$0.75$ with a relatively small gap with fully-supervised approaches.
The experimental results indicate that, despite utilizing only video-level annotations for weakly-supervised learning, the MDP could still effectively identify temporal forgery traces present within the multimodal features, enabling relatively precise localization of timestamps for forged segments.
This is achieved by analyzing the temporal deviations in the multimodal features, demonstrating the MDP's performance in addressing temporal forgery localization.

\textit{\textbf{AV-Deepfake1M Dataset:}}
Compared to LAV-DF, AV-Deepfake1M contains more Deepfake videos with long duration.
The longest duration of Deepfake video in LAV-DF is $19.97$s, while the longest video in AV-Deepfake1M is $32.51$s, and $4\%$ samples of the AV-Deepfake1M have a duration of more than $20$s.
Long duration videos are more challenging for weakly-supervised temporal forgery localization task that depend merely on video-level annotations.
Since the comparison weakly-supervised approaches 
could not localize the timestamps of forged segments in AV-Deepfake1M effectively, we have chosen the fully-supervised approaches (ActionFormer, TriDet, UMMAFormer and MFMS) to compare with the MDP on this dataset.

The experimental results are shown in Table \ref{av1m_comp}.
It could be observed that both fully-supervised approaches and weakly-supervised approach MDP have decreased in AR evaluation metric compared to the experimental results of LAV-DF.
As for AP, the proposed MDP underperforms on AP evaluation metrics with large IoU thresholds, while it achieves localization results close to the fully-supervised UMMAFormer on AP@$0.1$ and AP@$0.2$.
The experimental results show that MDP could mine temporal forgery traces and localize the timestamps of forged segments even in challenging AV-Deepfake1M dataset by relying merely on the video-level annotations.
We also conducted experiments with WS-TAL approaches CoLA and FuSTAL on the AV-Deepfake1M dataset.
The corresponding experimental results were not displayed in Table ~\ref{av1m_comp} since the obtained localization results are not effective.
The comparison results on this dataset also illustrate that the MDP could exploit the subtle forgery traces within the temporal features, thereby addressing the task of timestamps localization of the forged segments relying on the video-level annotations.
\begin{figure*}[htbp]
\centering
\includegraphics[width=0.78\textwidth]{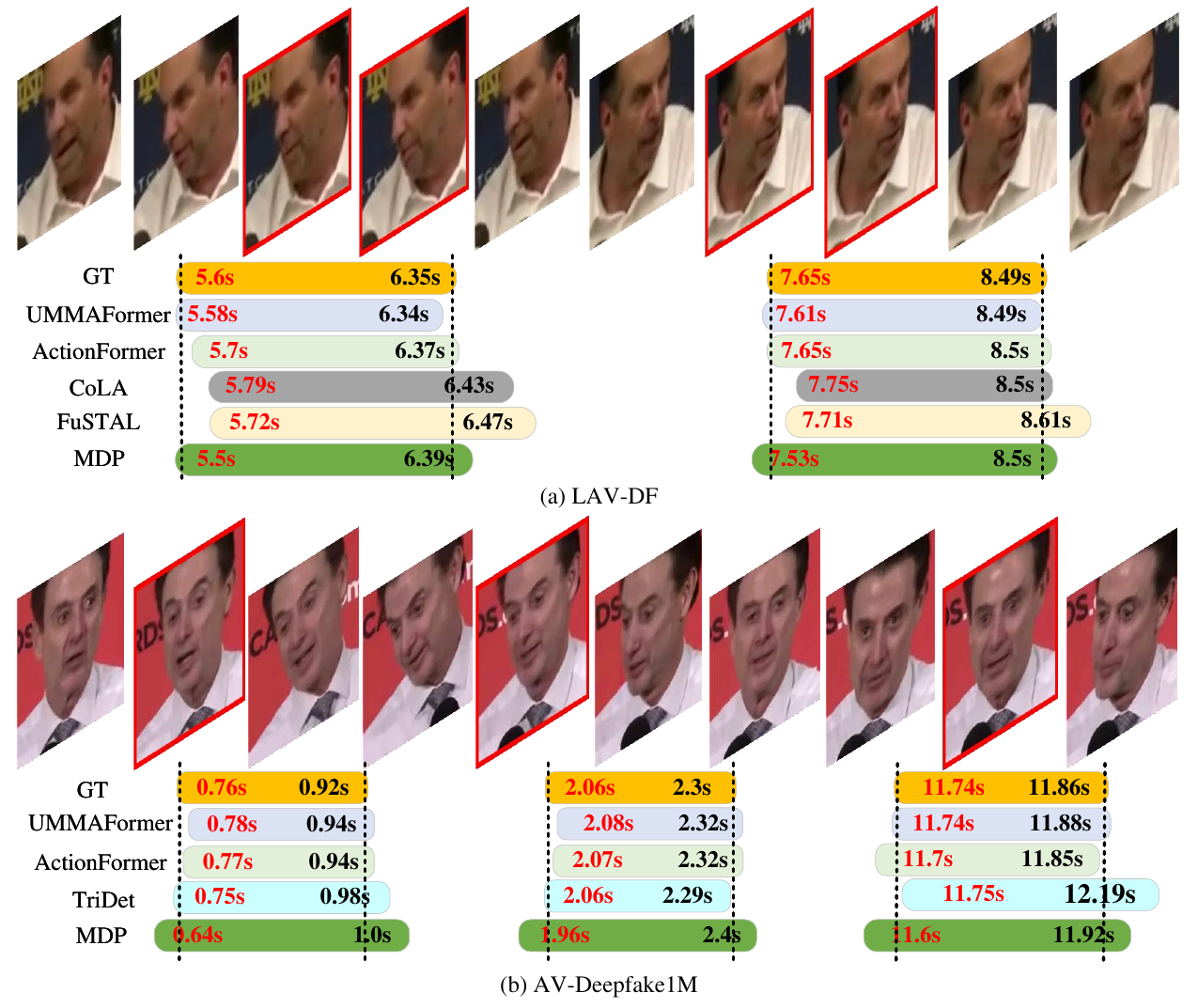}
\caption{Visualization results on the LAV-DF and AV-Deepfake1M. For LAV-DF, two fully-supervised approaches UMMAFormer, ActionFormer and two weakly-supervised approaches CoLA, FuSTAL are selected for visualization comparison. For AV-Deepfake1M, three fully-supervised approaches are selected for comparison. The red and black numbers indicate the start and end timestamps of the forged segments, respectively.}
\label{fig:lavdf_visualization}
\end{figure*} 
\begin{table}[h]
\centering
\setlength{\tabcolsep}{2pt} 
\begin{tabularx}{0.5\textwidth}{XX|XXXX|XXXXX}
\hline \multirow{2}{*}{CMA} & \multirow{2}{*}{$\mathcal{L}_{dp}$} & \multicolumn{4}{c|}{ AP@IoU(\%) } & \multicolumn{5}{c}{ AR@Proposals(\%) } \\
& & 0.5 & 0.75 & 0.95 & Avg. & 20 & 10 & 5 & 2 & Avg. \\
\hline
   &  & 72.47 & 44.4 & 0.05 & 38.97 & 61.69 & 61.69 & 61.5 & 59.66 & 61.14 \\
 $\checkmark$ & & 84.68 & 74.97 & 0.43 & 53.36 & 71.32 & 71.32 & 71.27 & 69.85 & 70.94 \\
   & $\checkmark$ & 85.64 & 66.25 & 0.16 & 50.68 & 67.6 & 67.29 & 64.82 & 52.51 & 63.03 \\
 $\checkmark$ & $\checkmark$ & 84.57 & 75.91 & 0.58 & $\mathbf{53.69}$ & 72.85 & 72.85 & 72.63 & 69.05 & $\mathbf{71.85}$ \\
 \hline
\end{tabularx}
	\caption{Ablation study about cross-modal attention (CMA) and deviation perceiving loss $\mathcal{L}_{dp}$. The experimental results of the ablation study are obtained on the LAV-DF dataset. The best average AP and AR are in bold.} 
	\label{ablation}
\end{table}
\subsection{Ablation Study}
This section conducts ablation studies on cross-modal attention and deviation perceiving loss.
In order to facilitate the interaction of visual and audio modality features, the MDP proposes a cross-modal attention with temporal property preservation based on feature alignment.
Additionally, since WS-TFL only has access to video-level annotations for loss calculation and model parameter learning, accurately localizing the timestamps of forged segments poses a significant challenge. 
The MDP introduces a deviation perceiving loss designed to help the model identify the temporal deviation of adjacent segments.

Comprehensive ablation studies are conducted on the LAV-DF dataset to further explore the effectiveness of the proposed components in MDP.
The results of the ablation study are shown in Table \ref{ablation}.
Specifically, we conducted four experiments.
The baseline is to generate the FAS $\mathcal{P}$ and video-level prediction result $\hat{y}$ by directly concatenating visual and audio features together after aligning them.
The other three experiments verify the temporal forgery localization performance after introducing CMA, $\mathcal{L}_{dp}$ and (CMA $+ \mathcal{L}_{dp}$), respectively.

It could be observed that the localization performance is significantly improved on both average AP ($+14.39\%$) and average AR ($+9.8\%$) with the enhancement of the cross-modal attention mechanism.
In multimodal Deepfake scenarios, the visual and audio modalities are typically embedded with extensive features, which are critical for mining
forgery traces.
In the spatial domain, the visual modality contains richer information compared to the audio modality.
Conversely,
In the temporal domain, the audio modality possesses a higher frequency of sampling points per unit of time than the visual modality.
Therefore, the proposed MDP aligns the visual and audio features in spatial and temporal domains.
Both visual and audio features are transformed into 1-D feature vectors on the spatial domain.
In the temporal domain, they are scaled to a uniform time dimension through a pooling operation.
This ensures the effective integration of multimodal data for better analysis.
For WS-TFL, the temporal information is apparently essential for the temporal forgery localization.
Subsequently, the cross-modal attention is utilized to enhance the audio features and visual features in a temporal property preservation manner, respectively.
The experimental results of the ablation study further validate the effectiveness of the proposed cross-modal attention component.

In addition, compared to the baseline, the introduction of deviation perceiving loss $\mathcal{L}_{dp}$ also improves the performance of temporal forgery localization on both average AP ($+11.71\%$) and average AR ($+1.89\%$).
Because WS-TFL could merely utilize the video-level annotations for model parameter learning, there is no temporal information to guide the model training, and locating the timestamps of forged segments is challenging in this background.
The previous Deepfake detection approaches have demonstrated that forged samples tend to have a larger deviation between adjacent segments compared to genuine samples.
The $\mathcal{L}_{dp}$ is based on measuring the deviation between adjacent segments, which consequently constrains the MDP to enlarge the temporal deviation of forgery samples while reducing that of genuine samples. 
The results of the ablation study indicate that $\mathcal{L}_{dp}$ could guide the MDP in perceiving the deviation between adjacent segments, and thus mine more temporal information for temporal forgery localization.
According to Table \ref{ablation}, the best average AP and average AR are achieved by introducing both CMA and $\mathcal{L}_{dp}$.
It validates the effectiveness of the key components in MDP.
\subsection{Visualization Analysis}
In order to display the performance of the MDP in temporal forgery localization, this section visualizes the localization results of MDP and comparison approaches.
The visualization results are shown in Figure ~\ref{fig:lavdf_visualization}.

From the visualization results, it could be observed that MDP could localize the timestamps of all the forged segments more precisely compared to the CoLA and FuSTAL on LAV-DF.
Moreover, the localization effectiveness of MDP is comparable to that observed in fully-supervised approaches within the presented sample.
In particular, the AV-Deepfake1M sample presents a notable challenge, as the duration of all three forged segments is below $0.3$s, while the overall duration of the Deepfake video is $23.36$s.
The localization of a small proportion of temporal forgeries is a significant challenge for WS-TFL.
Despite this, the MDP effectively utilizes weakly-supervised learning based solely on video-level annotations. 
It accurately predicts the timestamps of the forged segments, achieving comparable localization results to that obtained through fully-supervised approaches training at the frame-level annotations.
This represents that the proposed MDP effectively leverages the temporal forgery traces following the interaction of multimodal features, thereby enabling the precise identification of both the start and end timestamps of the forged segments.
\begin{table}[h]
	\centering
     \setlength{\tabcolsep}{2pt} 
\begin{tabularx}{0.5\textwidth}{c|XXXX|XXXXX}
 \hline \multirow{2}{*}{$f(\cdot)$} & \multicolumn{4}{c|}{ AP@IoU(\%) } & \multicolumn{5}{c}{ AR@Proposals(\%) } \\
 & 0.5 & 0.75 & 0.95 & Avg. & 20 & 10 & 5 & 2 & Avg. \\
\hline
   $L_{1}$ & 82.72 & 65.06 & 0.09 & 49.29 & 64.33 & 61.77 & 54.15 & 34.62 & 53.72 \\
   $L_{2}$& 85.04 & 72.05 & 0.59 & 52.56 & 70.3 & 70.24 & 69.76 & 66.56 & 69.22 \\
    MSE & 84.57 & 75.91 & 0.58 & $\mathbf{53.69}$ & 72.85 & 72.85 & 72.63 & 69.05 & $\mathbf{71.85}$ \\
 \hline
\end{tabularx}
	\caption{Temporal forgery localization results of different deviation measure functions $f(\cdot)$. The experimental results are obtained on the LAV-DF dataset.  The best average AP and AR are in bold.} 
	\label{dpf}
\end{table}
\subsection{Deviation Measure Function}
\label{DMF}
While calculating the $\mathcal{L}_{dp}$, a deviation measure function $f(\cdot)$ is required to measure the deviation between adjacent segments.
In this section, several experiments are conducted to test the effectiveness of different $f(\cdot)$ on the performance of temporal forgery localization.

Considering the computational complexity and parallelism, three deviation perceiving methods, Manhattan distance ($L_{1}$), Euclidean distance ($L_{2}$) and mean square error (MSE), are selected for the experiments.
The experimental results are shown in Table \ref{dpf}. 
It could be observed that the selection of $f(\cdot)$ has an obvious influence on the performance of MDP.
MSE achieves the best performance among the three measure methods.
In addition, compared to the baseline in Table \ref{ablation}, the localization performance improves after introducing the $\mathcal{L}_{dp}$ based on $L_{2}$ and MSE.
This further illustrates the value of the deviation perceiving idea proposed in MDP for WS-TFL task.
It should be noted that the deviation measure function discussed in this paper remains an open problem.
Investigating more effective $f(\cdot)$ represents a meaningful research direction.

\section{Conclusion}
In this paper, we propose a multimodal deviation perceiving framework for weakly-supervised temporal forgery localization (MDP), which aims to localize the start and end timestamps relying merely on video-level annotations.
The MDP presents an innovative multimodal interaction mechanism that focuses on the alignment of multimodal features, involving cross-modal attention to dig inter-modality deviation between visual and audio features while preserving the temporal property.
Moreover, an extensible deviation perceiving loss is introduced to enlarge the temporal deviation of adjacent segments of the forged samples and reduce that of genuine samples.
The experiments conducted on two challenging datasets, LAV-DF and AV-Deepfake1M, demonstrate the effectiveness of the MDP.
The localization performance of MDP is close to the fully-supervised approaches in some evaluation metrics.
In the future, the proposed framework requires further improvements to enhance the precision of timestamp localization of forged segments.
Weakly-supervised temporal forgery localization (WS-TFL) based on multimodal deviation perceiving deserves to be further explored, especially in multimodal feature interaction and deviation measure function.

\clearpage
\begin{acks}
This work is supported by the National Natural Science Foundation of China (No.62441237, No.62172435).
\end{acks}

\bibliographystyle{ACM-Reference-Format}
\bibliography{reference}










\end{document}